\documentclass[sigconf,table]{acmart}

\usepackage[english]{babel}
\usepackage{type1ec}
\usepackage[utf8]{inputenc}
\usepackage[T1]{fontenc}
\usepackage{subcaption,graphicx}
\usepackage{pbox}
\usepackage{makecell}

\usepackage{amssymb}
\usepackage{amsthm}
\usepackage{multirow}
\usepackage{fancyvrb}
\usepackage{algpseudocode}
\usepackage{algorithm}
\usepackage{algpseudocode}
\usepackage{subcaption}
\usepackage{mathtools}
\usepackage{qtree}
\usepackage{alltt}
\usepackage{lipsum}  
\usepackage{makecell}
\usepackage{tablefootnote}
\usepackage{bibentry}
\usepackage{geometry}
\usepackage{booktabs}
\usepackage{color, colortbl}
\usepackage{hyperref}

\definecolor{officegreen}{rgb}{0.0, 0.5, 0.0}
\definecolor{byzantium}{rgb}{0.44, 0.16, 0.39}
\definecolor{darkturquoise}{rgb}{0.0, 0.81, 0.82}

\setcopyright{rightsretained}

\copyrightyear{2020}
\acmYear{2020}
\setcopyright{acmcopyright}\acmConference[GECCO '20]{Genetic and Evolutionary Computation Conference}{July 8--12, 2020}{Cancún, Mexico}
\acmBooktitle{Genetic and Evolutionary Computation Conference (GECCO '20), July 8--12, 2020, Cancún, Mexico}
\acmPrice{15.00}
\acmDOI{}
\acmISBN{}

\begin{document}
\title{A Robust Experimental Evaluation of Automated Multi-Label Classification Methods}

\author{Alex G. C. de S\'a}
\affiliation{%
  \institution{Computer Science Department \\ Universidade~Federal~de~Minas~Gerais}
  \city{Belo Horizonte, MG} 
  \state{Brazil} 
}
\email{alexgcsa@dcc.ufmg.br}

\author{Cristiano G. Pimenta}
\affiliation{%
  \institution{Computer Science Department \\ Universidade~Federal~de~Minas~Gerais}
  \city{Belo Horizonte, MG} 
  \state{Brazil} 
}
\email{cgpimenta@dcc.ufmg.br}

\author{Gisele L. Pappa}
\affiliation{%
  \institution{Computer Science Department \\ Universidade~Federal~de~Minas~Gerais}
  \city{Belo Horizonte, MG} 
  \state{Brazil} 
}
\email{glpappa@dcc.ufmg.br}

\author{Alex A. Freitas}
\affiliation{
  \institution{School of Computing \\ University of Kent}
  \city{Canterbury,~Kent} 
  \state{United Kingdom} 
}
\email{A.A.Freitas@kent.ac.uk}

\renewcommand{\shortauthors}{}

\begin{abstract}
Automated Machine Learning (AutoML) has emerged to deal with the selection and configuration of algorithms for a given learning task. With the progression of AutoML, several effective methods were introduced, especially for traditional classification and regression problems. Apart from the AutoML success, several issues remain open. One issue, in particular, is the lack of ability of AutoML methods to deal with different types of data. Based on this scenario, this paper approaches AutoML for multi-label classification (MLC) problems. In MLC, each example can be simultaneously associated to several class labels, unlike the standard classification task, where an example is associated to just one class label. In this work, we provide a general comparison of five automated multi-label classification methods -- two evolutionary methods, one Bayesian optimization method, one random search and one greedy search -- on 14 datasets and three designed search spaces. Overall, we observe that the most prominent method is the one based on a canonical grammar-based genetic programming (GGP) search method, namely Auto-MEKA$_{GGP}$. Auto-MEKA$_{GGP}$ presented the best average results in our comparison and was statistically better than all the other methods in different search spaces and evaluated measures, except when compared to the greedy search method. 
\end{abstract}



\maketitle

\section{Introduction}

We are experiencing the era of data. With its great availability, people in general (e.g., practitioners, data scientists, and researchers) are trying hard to extract useful information encoded on data~\cite{Siegel2013}. This resulted in an ever-growing popularity and the indiscriminate use of machine learning (ML) algorithms by many types of users.

The field of \emph{Automated Machine Learning}~(AutoML) \cite{Hutter2019} has emerged to help this wide and heterogeneous public.  This field has the purpose of democratizing ML in a way ML can be used with less difficulties by general audiences. In addition, AutoML also aims to assist experienced data scientists.  In both scenarios, the field of AutoML has the scope of recommending learning algorithms (and often their hyper-parameters' settings too)  when people face a particular problem that might be (partially or totally) solved with ML. Broadly speaking, AutoML proposes to deal with users' biases by customizing the solutions (in terms of algorithms and configurations) to ML problems following different approaches.

AutoML has been successfully and mainly employed to solve traditional (single-label) classification and regression problems~\cite{Elshawi2019}. However, this work is interested in AutoML methods for a different and specific type of data, called \emph{Multi-Label Classification}~(MLC)~\cite{Tsoumakas2011,Zhang2014,Herrera2016}. The goal of MLC is to learn a model that expresses the relationships between a set of predictive features (attributes) describing the examples and a predefined set of class labels. In MLC, each example can be simultaneously associated to one or more class labels. Each class label is represented by a discrete value. 

When compared to single-label classification (SLC), MLC can be considered a more challenging task, mainly due to the following reasons. First, an MLC algorithm needs to consider the label correlations (i.e., detecting whether or not they exist) in order to learn a model that produces accurate classification results ~\cite{Zhang2014}. Second, given the usual limited number of examples for each class label in the dataset~\cite{Herrera2016}, the generalization in MLC is considered harder than SLC, as the MLC algorithm needs more examples to create a good model from such complex data~\cite{Domingos2012}. Third, there is a strain to evaluate MLC classifiers as several metrics follow contrasting aspects to define what is a good MLC prediction~\cite{Pereira2018}. Finally, the learning algorithms applied to solve MLC problems need more computational resources than the ones used to solve SLC~\citep{Herrera2016}. This is mainly due to MLC being a generalization of SLC, so that the algorithms need to look at several labels instead of just one. 

We claim that these aforementioned challenges are part of the reason why AutoML for MLC problems (i.e., AutoMLC) has not been sufficiently explored. Taking it into consideration, this work performs an assessment of popular \emph{search methods} for AutoMLC, including evolutionary methods, a Bayesian optimization method and blind-\emph{search methods}. For that, we propose two novel AutoMLC \emph{search methods}. The first is an extension of the work of de Sá \emph{et al.}~\cite{deSa2018} on Grammar-based Genetic Programming (GGP) \cite{McKay2010,Whigham1995} for AutoMLC, named Auto-MEKA$_{GGP}$. Our extension adds to the GGP core a speciation approach~\citep{Back1999} aiming to improve the diversity of the produced solutions. The second \emph{search method} is a Bayesian optimization (BO) method, namely Sequential Model-based Algorithm Configuration~(SMAC) \cite{Hutter2011} -- note that there was no such methods previously proposed in the AutoMLC literature. As both proposed methods are based on the well-known MLC  MEKA framework \cite{Read2016}, we named these \emph{search methods} as Auto-MEKA$_{spGGP}$ and Auto-MEKA$_{BO}$, respectively.

We compare these two proposed \emph{search methods} with Auto-MEKA$_{GGP}$~\cite{deSa2018}, a random search (namely, Auto-MEKA$_{RS}$) and a greedy search (namely, Auto-MEKA$_{GS}$) on three designed MLC search spaces (namely, {Small}, {Medium} and {Large}) over 14 benchmarking datasets. Finally, in this work, we use five performance measures for evaluating these methods, due to the additional degree of freedom that the MLC algorithms' setting introduces~\citep{Madjarov2012}.

The experimental results show that Auto-MEKA$_{GGP}$ mostly presented the best average results and also the best average ranks for several \emph{search spaces} and measures. Besides, Auto-MEKA$_{GGP}$ was the only method to be statistically better than all other evaluated \emph{search methods} in different occasions (i.e., performance measures \emph{versus} \emph{search spaces}), except when compared to Auto-MEKA$_{GS}$.

Although this is a positive result for the evolutionary methods, we believe that more robust methods -- such as Auto-MEKA$_{GGP}$, Auto-MEKA$_{spGGP}$ and Auto-MEKA$_{BO}$-- can still improve their predictive performances. With this in mind, we observe that these methods could not satisfactorily trade-off between exploration and exploitation as they were not statistically and simultaneously better than pure-exploration and pure-exploitation methods (i.e., Auto-MEKA$_{RS}$ and Auto-MEKA$_{GS}$, respectively).

The results also show that there is a high correlation between the size (and definition) of the search space and the effectiveness of AutoMLC methods to select and configure algorithms. When looking at the predictive accuracy of the AutoMLC methods, we have an indication that as the size of an AutoMLC's \emph{search space} decreases, pure-exploration and/or pure-exploitation AutoMLC \emph{search methods} tend to have similar results to robust AutoMLC methods (such as the ones presented in this work).

The remainder of this paper is organized as follows. Section~\ref{sec:theoretical} introduces MLC and Section \ref{sec:automlc} reviews related work on AutoMLC. Section~\ref{sec:methods} details AutoMLC in terms of the proposed \emph{search spaces} and evaluated \emph{search methods} that are included in the experimental comparison, while Section~\ref{sec:experiments} presents and discusses the obtained results. Finally, Section~\ref{sec:conclusions} draws some conclusions and discusses directions for future work.
\section{Multi-label classification}
\label{sec:theoretical}

There is a great number of works on traditional single-label classification (SLC) for machine learning (ML)~\cite{Zaki2020}. In SLC, each example is defined by a tuple ($X, y$), where ${X} = \{x_1, ..., x_d\}$ is a $d$-dimensional vector representing the feature space (i.e., the categorical and/or numerical characteristics of that example) and $y$ is the class value, where  $y \in L$, a set of disjoint class labels. In SLC, each example is strictly associated to a single class label.

Nevertheless, there is an increasing number of applications that require associating an example to more than one class label \cite{Gibaja2015}, such as medical diagnosis and protein function prediction. This classification scenario is better known as \emph{Multi-Label Classification}~(MLC). According to \cite{Tsoumakas2010}, each example in MLC is represented by a tuple  (${X}$, ${Y}$), where ${X}$ is the $d-$dimensional feature vector, and ${Y} \subseteq L$ is a set of non-disjoint class labels. Hence, we would like to find an MLC model $h$: ${X} \rightarrow 2^{|L|} $ such that $h$ maximizes a quality criterion $\lambda$. 

The literature divides MLC algorithms into three categories \citep{Gibaja2015}: problem transformation~(PT), algorithm adaptation~(AA) and ensemble or meta-algorithms methods (Meta-MLC). Whereas PT methods transform the multi-label problem into one or more single-label classification problems, AA methods simply extend single-label classification algorithms so they can directly handle multi-label data. Finally, Meta-MLC methods act on top of PT or AA multi-label classifiers, aiming to combine the results of MLC algorithms and produce models with more robust predictive performances. 

Among the great number of MLC algorithms~\cite{Tsoumakas2011,Zhang2014,Herrera2016}, it is important to mention three methods that transform an MLC problem into one or many SLC problems: Label Powerset (LP), Binary Relevance (BR) and Classifier Chain (CC). LP creates a single class for each unique set of labels that is associated with at least one example in a multi-label training set.  BR, in turn, learns $|L|$ independent binary classifiers, one for each label in the labelset $L$. Finally, CC changes the BR method by chaining the binary classifiers. In this case, the feature space of each link in the chain is increased with the classification outputs of all previous links. 

\section{Related Work on AutoMLC}
\label{sec:automlc}

Most AutoML methods in the literature were designed to solve the conventional single-label classification and regression tasks \cite{Elshawi2019}, and can not handle multi-label data. As far as we know, there are only a few works related to \emph{Automated Multi-Label Classification}~(AutoMLC). 

\cite{Chekina2011} developed a meta-model (i.e., a $20$-Nearest Neighbors classifier) for selecting one out of 11 multi-label classification algorithms, taking into account 30 characterizing measures and 36 meta-datasets. Nevertheless, as it is a preliminary work, it only selects the MLC algorithm, not setting the algorithm's hyper-parameters.

Evolutionary Multi-Label Ensemble (EME) \citep{Moyano2019} encompasses the problem of selecting MLC algorithms to compose MLC ensembles. The main idea of EME stands on the simplicity of each ensemble's multi-label classifier, which is focused on a small subset of the labels, but still considering the relationships among them and avoiding the high complexity of the output space. Nevertheless, EME takes into account only one type of model to compose the ensembles (i.e., the model produced by label powerset), so it is not sufficient to deal with all types of MLC problems.

Furthermore, \cite{Wever2019} proposed an extension to a canonical hierarchical planing method (i.e., ML-Plan) to the MLC context. They named this method as ML$^2$-Plan (Multi-Label ML-Plan). Basically, ML$^2$-Plan is implemented as a global best-first search over the graph induced by the planning problem at hand.

Finally, \cite{deSa2017,deSa2018} proposed two AutoML methods for MLC problems: GA-Auto-MLC and Auto-MEKA$_{GGP}$. Whereas GA-Auto-MLC employs a real-coded genetic algorithm \cite{Eiben2003} to perform its search, Auto-MEKA$_{GGP}$ uses a grammar-based genetic programming algorithm~\cite{McKay2010,Whigham1995}. Auto-MEKA$_{GGP}$ is a robust enhancement of GA-Auto-MLC to handle huge and, consequently, complex MLC \emph{search spaces}. Because of that, we decided to only include  Auto-MEKA$_{GGP}$ into our experiments. We will further discuss it in Section \ref{sec:methods}.
\section{AutoML Methods for Multi-label Classification}
\label{sec:methods}

\begin{figure}[!tb]
  \centering
   \includegraphics[scale=0.2605]{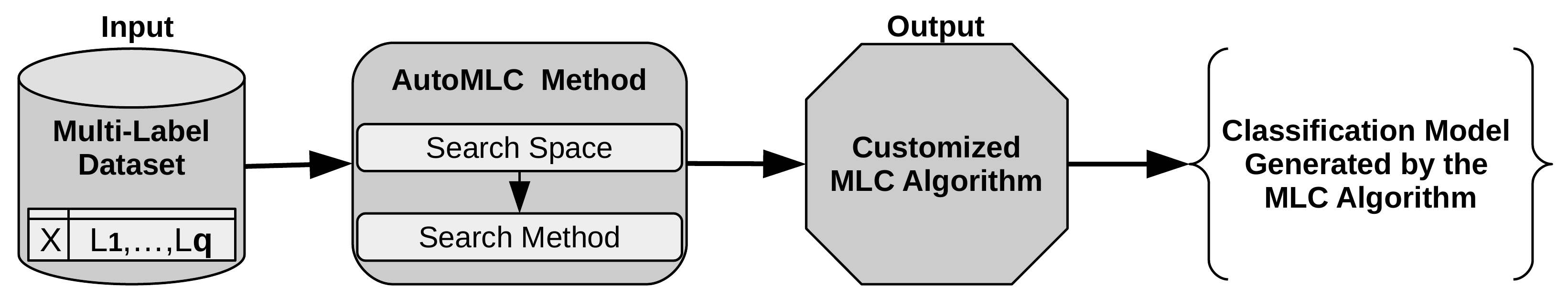}
   \caption{The general AutoMLC framework to select and configure MLC algorithms.}
   \label{fig-general}
 \end{figure}

This section introduces a generic AutoMLC framework followed by all AutoMLC methods evaluated in this paper. As illustrated in Figure~\ref{fig-general}, the AutoMLC method receives as input a specific multi-label dataset (with the feature space $X$ and the class labels $L_1$ to $L_q$). Structurally, the evaluated AutoMLC methods have two main components: the \emph{search space} and the \emph{search method}. 

The \emph{search space} consists of the main building blocks (e.g., the prediction threshold values, the hyper-parameters and the algorithms at the SLC level) from previously designed MLC algorithms. To explore this \emph{search space}, the AutoMLC method uses a \emph{search method}, which finds appropriate MLC algorithms to the dataset at hand. However, the performance of the \emph{search method} depends on what is specified in the \emph{search space}. 

Once the search is over, the AutoMLC method outputs an MLC algorithm tailored to the input dataset based on that \emph{search space}. This MLC algorithm is specifically selected and (hyper-)parameterized to this dataset, although it could be applied to any multi-label dataset. In the end, the customized MLC algorithm returns an MLC model, and consequently, its classification results.

The evaluation presented in this paper considers three \emph{search spaces}, namely Small, Medium and Large, which differ from each other in terms of complexity. These three \emph{search spaces} are explored using five \emph{search methods}: Auto-MEKA$_{GGP}$, Auto-MEKA$_{SpGGP}$, Auto-MEKA$_{BO}$, Auto-MEKA$_{RS}$ and Auto-MEKA$_{GS}$, as detailed in Section~\ref{search-methods}.

\subsection{Search Spaces} \label{search-space_auto-ml}

To design the \emph{search spaces} for the AutoMLC methods being evaluated, we first performed a deep study about multi-label classification in the MEKA software. We analyzed in detail all the algorithms and their hyper-parameters, the constraints associated with different hyper-parameter settings, the hierarchical nature of operations performed by problem transformation methods and meta-algorithms, among other issues.

Based on that, we designed three \emph{search spaces}\footnote{\label{suppl_mat}For more details about the MLC and SLC algorithms and meta-algorithms that compose the \emph{search spaces}, see the supplementary material.}: Small, Medium and Large. The reason behind this threefold modeling is basically because we want to test different levels of \emph{search space} complexity. 

For the \emph{search space} Small, for instance, we have five MLC algorithms, where four of them can be combined with other five SLC algorithms, as they are from the PT category. The only algorithm that can not be combined with SLC algorithms is ML-BPNN,  which belongs to the AA category. Therefore, the \emph{search space} Small consists of 10 learning algorithms -- five MLC algorithms and five SLC algorithms, which gives a set of 21 combinations of learning algorithms,  where the AA category counts as one combination.

In contrast, the \emph{search space} Medium has 30 learning algorithms -- 15 MLC algorithms and 15 SLC algorithms, which produces 211 combinations of algorithms. Finally, the \emph{search space} Large has a total of 54 learning algorithms -- 26 MLC algorithms and 28 SLC algorithms, which produces 16,568 possible combinations of learning algorithms.

Although the main difference between the Small and Medium \emph{search spaces} is the number of learning algorithms, note that, when comparing Medium to Large, we have a change on the structure of the \emph{search space}. This happens because we only added meta-algorithms at the MLC and SLC levels into Large. Hence, we have more levels in the multi-label hierarchy to consider. For example, when a \emph{search method} is selecting a new MLC algorithm in this \emph{search space}, it must decide whether to include or exclude meta-algorithms at the MLC and SLC levels. As a result, this \emph{search space} is hierarchically more complex than the other two (i.e., Small and Medium).

Taking into account the number of learning algorithms, the number of hyper-parameters, and the constraints in the choices of algorithms' components and (hyper)-parameters in MEKA, we estimated the size of the three \emph{search spaces}\footnote{In these estimations, the real-valued hyper-parameters have always taken 100 different discrete values.}. In total, the \emph{search space} Small has $[(5.070 \times 10^{7}) + (8.078 \times 10^8 \times m)  +  (2.535 \times 10^{10} \times q)]$ possible MLC algorithm configurations (i.e., a given set of learning algorithms with their respective hyper-parameters), where $m$ is the number of features (or attributes) and $q$ is the number of labels of the dataset. The \emph{search space} Medium, on the other hand, is estimated as having $[(2.545 \times 10^{16}) + (8.078 \times 10^{8} \times m)  + (1.151 \times 10^{12} \times q)]$ possible MLC algorithm configurations. Finally, the \emph{search space} Large is estimated to have approximately $[(6.555 \times 10^{29}) + (1.443 \times 10^{14} \times m) + (3.042 \times 10^{22} \times q) + (1.291 \times 10^{27} \times \sqrt{q})]$ possible MLC algorithm configurations.

\subsection{Search Methods} \label{search-methods}

This section details the five \emph{search methods} used in our comparison. They all follow the same methodology.

Each \emph{search method} starts its own iterative process by generating, evaluating and looking for MLC algorithms configurations. To perform the evaluation, the \emph{search methods} use the average of four well-known measures \cite{Tsoumakas2010,Gibaja2015}: Exact Match~(EM), Hamming Loss~(HL), $F_1$ Macro averaged by label~(FM) and Ranking Loss~(RL), as indicated in Equation \ref{eq}. The \emph{search method} keeps iterating while a maximum time budget is not reached. At the end, the best MLC algorithm configuration in accordance to the quality criteria is returned and assessed in the test set.

\begin{equation}\label{eq}
  Fitness = \frac{EM + (1 - HL) + FM + (1 - RL)}{4}
\end{equation}

Regarding the MLC evaluation measures, EM is a very strict  metric, as it only takes the value one when the predicted label set is an exact match to the true label set for an example, and takes the value zero otherwise. HL, in turn, calculates how many example-label pairs are misclassified. Furthermore, FM is the harmonic mean between precision and recall, and its average is first calculated for each label and, after that, across all the labels. Finally, RL measures the number of times that irrelevant labels are ranked higher than relevant labels, i.e., it penalizes the label pairs that are reversely ordered in the ranking for a given example. All four metrics are within the $[0,1]$ interval. However, whereas the EM and FM measures should be maximized, the HL and RL measures should be minimized. Hence, HL and RL are subtracted from one in Equation~\ref{eq} to make the search maximize the fitness function. 

\subsubsection{$\text{Auto-MEKA}_{\text{GGP}}$}

This method was proposed in \cite{deSa2018}, and relies on a Grammar-based Genetic Programming (GGP) approach~\cite{McKay2010, Whigham1995}, which has the advantage of hierarchically exploring the nature of the AutoMLC problem. In this case, the grammar encompasses the \emph{search space} of MLC algorithms and hyper-parameter settings.

In $\text{Auto-MEKA}_{\text{GGP}}$, each individual expresses an MLC algorithm configuration, and is represented by a derivation tree generated from the grammar. Individuals are first generated by choosing at random a valid production rule, and then mapping it into an MLC algorithm (with a specific hyper-parameter setting).

\begin{figure}[!th]
   \centering
   \includegraphics[scale=0.225]{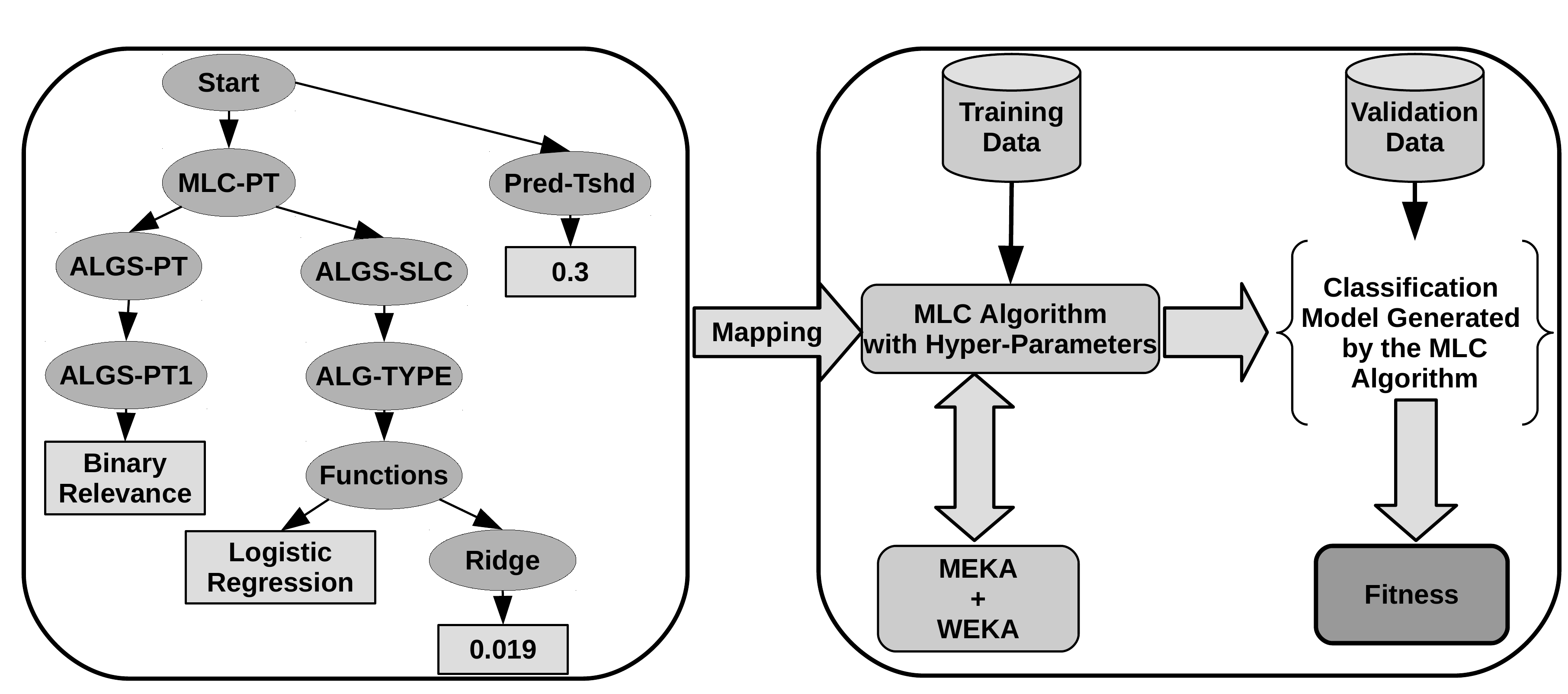}
   \caption{Individual's evaluation process in Auto-MEKA$_{GGP}$.}
    \label{figEvaluation}
  \end{figure}

Figure \ref{figEvaluation} details the whole mapping process followed by the evaluation process for each GGP individual. In the example of Figure~\ref{figEvaluation}, ellipsoid nodes are the grammar's non-terminals, whereas the rectangles are the terminals.

The mapping process takes the terminals from the tree and constructs a valid MLC algorithm. The mapping in Figure~\ref{figEvaluation} will produce the following MLC algorithm: a Binary Relevance method combined with a Logistic Regression algorithm (with the hyper-parameter ridge set to 0.019), using a threshold of 0.3 to classify the MLC data. 

Next, individuals have their fitness calculated as previously explained, and undergo tournament selection. The GGP operators (i.e., Whigham's crossover and mutation~\citep{Whigham1995}) are applied to the selected individuals to create a new population. These operators also respect the grammar constraints, ensuring that the produced individuals represent valid solutions.

\subsubsection{$\text{Auto-MEKA}_{\text{spGGP}}$}

This novel AutoMLC method enhances the search mechanisms of Auto-MEKA$_{GGP}$ by adding a speciation process~\citep{Back1999} into its \emph{search method}. The general idea is to use Grammar-based Genetic Programming with Speciation (spGGP) to improve the trade-off between exploration and exploitation of the search for MLC algorithms and hyper-parameter settings. Because the proposed \emph{search spaces} have an exponential size and a complex hierarchical nature, it may be crucial to use this approach to deal with these aspects. A species is a set of individuals that resemble each other more inherently than the individuals in another species~\citep{Back1999}. In speciation-based evolutionary computation, the objective is to emphatically restrict mating to those among like individuals from the population. In this case, likeness among individuals is identified if they have similar genotypes or phenotypes.  

In this work, we defined a set of species based on the types of MLC hyper-parameters (i.e., categorical, discrete or continuous) and their interactions. It is worth noting that, whilst the categorical hyper-parameters take Boolean and string-based values (e.g., an algorithm name or a procedure name in an algorithm), the discrete hyper-parameters only take integer values. Therefore, our speciation-based method focuses not exclusively on the choice of the learning algorithms but primarily on the different types of hyper-parameters, where the choice of the learning algorithms is set as a special case of a categorical hyper-parameter. 

In general, we would like to understand if there is a dependence between the final AutoMLC predictive performance and the types (and the interactions) of hyper-parameters for a given dataset. For instance, if we would like to recommend MLC algorithms for two datasets with different characteristics, understanding only the categorical hyper-parameters for the first dataset may be more beneficial than understanding discrete and/or continuous hyper-parameters. This could be the opposite for the second dataset.

In this context, we design eight species. Different species specialize on optimizing different combinations of hyper-parameter types and their interactions. All species have instances of all learning algorithms at both MLC and SLC levels based on the defined \emph{search space}, but they vary on the types of hyper-parameters that are left with their default values during evolution and cannot be updated. In the descriptions below, the settings of the hyper-parameters we refer to can be changed by the evolutionary process, while all others are set to their default values. The species may vary:

\begin{enumerate}
    \item \textbf{Learning algorithms: }Only the categorical hyper-parame-ters referring to the names of the (traditional and meta) learning algorithms at the MLC and SLC levels can be combined and evolved.
    
    \item \textbf{Learning algorithms and common categorical hyper-parameters: }Together with the categorical hyper-parameters indicating the names of the learning algorithms (species 1), this species also allows the combination and evolution of common categorical hyper-parameters (e.g., the names of a metric). In addition, this species also encompasses Boolean hyper-parameters. 
    
    \item \textbf{Learning algorithms and discrete hyper-parameters: }This species considers, alongside with the categorical hyper-parameters that indicate the names of the learning algorithms, the discrete (integer) hyper-parameters. 
    
    \item \textbf{Learning algorithms and continuous hyper-parameters: }This species allows the modification and combination of the continuous hyper-parameters of species~1. 
  
    \item \textbf{Learning algorithms and the combination of common categorical and discrete hyper-parameters: }In this species, we evolve the individuals considering the learning algorithms themselves (species~1) together with common categorical and discrete hyper-parameters. 
  
    \item \textbf{Learning algorithms and the combination of common categorical and continuous hyper-parameters: }In this case, we make the evolutionary process consider the hyper-parameters representing the learning algorithms, the common categorical hyper-parameters and the continuous hyper-parameters. 
   
    \item \textbf{Learning algorithms and the combination of discrete and continuous hyper-parameters: }This species allows the combination of the names of the learning algorithms with discrete and continuous hyper-parameters. 
    \item \textbf{All types of hyper-parameters: }This species is more general, and all types of hyper-parameters (categorical referring to the names of the learning algorithms, common categorical, discrete, continuous hyper-parameters) are considered to be explored/exploited.

\end{enumerate}

The first step of $\text{Auto-MEKA}_{\text{spGGP}}$'s evolutionary process is the initialization procedure, where we generate for each species a population of individuals, which are represented by trees and built based on a specific grammar for that species.

$\text{Auto-MEKA}_{\text{spGGP}}$ also differs from $\text{Auto-MEKA}_{\text{GGP}}$ in the cross-over operator, which can be performed for both intra-species and inter-species individuals. By interchangeably using both types of crossover operations, we have more chances to test unknown regions of the \emph{search space} (exploration) when using the inter-species crossover, while a more local search over the different types of hyper-parameters is performed in each species by the intra-species crossover (exploitation).

It is worth noting that we decided to design the mutation operator as a local operator in each specie. By doing that, Whigham's mutation uses the \emph{grow} method on the individual's derivation tree but ensures that the MLC grammar of the current species is applied over the \emph{grow} method.

\subsubsection{$\text{Auto-MEKA}_{\text{BO}}$}

This proposed Bayesian Optimization (BO) AutoMLC method employs the Sequential Model-based Algorithm Configuration (SMAC) \cite{Hutter2011} as a procedure to search for suitable MLC configurations. In our generic framework, Auto-MEKA$_{BO}$ can be categorized as a sophisticated extension of Auto-WEKA~\cite{Thornton2013}.

Hence, given a dataset and a \emph{search space}, $\text{Auto-MEKA}_{\text{BO}}$ uses a performance model (in our case, a Random Forest) to robustly select the MLC configurations. This model is initialized with a default MLC algorithm with default hyper-parameter settings. In the case of $\text{Auto-MEKA}_{\text{BO}}$, we initialize the model with different algorithms as the \emph{search spaces} allow different types of learning algorithms. For the \emph{search space} Small, we run and include into the model the results of the classifier chain~(CC) algorithm using the na\"ive Bayes~(NB) classification algorithm at the single-label base level.

As the \emph{search space} Medium is similar to Small in terms of the hierarchical levels, we keep the CC algorithm at the multi-label level. However, we have tried to improve the single-label classification level by using a strong algorithm, i.e., we use a more sophisticated Bayesian network classifier~(BNC) algorithm instead of a simple NB classification algorithm. Hence, at this level, the K2 algorithm is employed.

Finally, for the \emph{search space} Large, we define as the initial configuration to the model the random subspace meta-algorithm for multi-label classification~(RSML), using the Bayesian classifier chain~(BCC) algorithm at the multi-label base level, the locally weighted learning~(LWL) algorithm at the single-label meta level, and the BNC K2 algorithm at the single-label base level. Except for RSML, which is a robust meta-algorithm, the other levels were chosen in an arbitrary fashion, although they are also considered strong algorithms in the machine learning literature.

After this initialization step, we choose the next configuration from the MLC \emph{search space} in the configuration files, relying on this performance model. To do that, the SMAC method is used to select a better configuration. Next, this MLC configuration is evaluated in the MEKA framework and then compared with the best MLC configuration found so far. If the current configuration has a better score than the current best configuration, it is saved and set as the new best configuration. Otherwise, the process continues by verifying if the time budget was reached. If this criterion is met, $\text{Auto-MEKA}_{\text{BO}}$ returns the best configuration found up to now. Otherwise, the last evaluated MLC configuration is added to the performance model with its corresponding quality value, updating it. The process continues by following these same steps until the time budget expires. 

\subsubsection{Auto-MEKA$_{RS}$}

The AutoMLC random \emph{search method} (RS) iterates over the predefined MLC \emph{search space} at random. First, it creates $p$ MLC algorithm configurations, evaluates them and saves the best configuration in terms of the proposed quality measure (see Equation \ref{eq}) into a list. Next, it creates other $p$ new MLC algorithm configurations, evaluates these configurations and saves the best at this iteration into the same list. RS keeps doing this procedure until the total time budget is reached. At the end, it returns the best MLC algorithm configuration from the list based on the quality measure. 

\subsubsection{Auto-MEKA$_{GS}$}

The AutoMLC greedy \emph{search method} (GS) starts by generating an initial random solution (i.e., an MLC algorithm configuration), which is set as the current best. From this solution, we generate $p$ others by performing local changes into its representation. We use the aforementioned grammar-based representation for both random and greedy searches. Thus, from the grammar, we generate a derivation tree and employ Whigham's mutation \cite{McKay2010,Whigham1995} to perform local operations in the respective tree. We evaluate these solutions (see Equation \ref{eq})  and check if one of them has a better quality score than the current best MLC configuration. If so, we update the best configuration with the best score. Otherwise, we maintain the best MLC configuration. Next, from the current best configuration, we continue looking at its neighbors to create, evaluate and possibly find new better solutions. This search process remains while the final time budget is not reached. At the end, the best found MLC algorithm configuration is returned based on the proposed quality measure.
\section{Experimental Analysis}
\label{sec:experiments}

This section presents the experimental results of the AutoMLC methods discussed in the previous section. The experiments involve a set of 14 datasets selected from the KDIS (Knowledge and Discovery Systems) repository\footnote{The datasets are also available at: \url{http://www.uco.es/kdis/mllresources/}.}, as described in Table~\ref{table-datasets}. These datasets were chosen based on their differences in application domain, the number of instances ($n$), the number of features ($m$), the number of labels ($q$), the label cardinality -- the average number of labels associated with each example in the dataset (Card.), the label density -- the average number of labels associated with each example divided by the number of labels (Dens.), and the label diversity -- the percentage of labelsets in the dataset divided by the number of possible labelsets (Div.). 

\begin{table}[htbp]
\scriptsize
\caption{Datasets used in the experiments.}
\begin{center}
\begin{tabular}{|l||c|c|c|c|c|c|c|c|}
\hline
\textbf{Datasets}                 & \textbf{Acronym}   &  $\textbf{n}$   & $\textbf{m}$     & $\textbf{q}$   & \textbf{Card.}     & \textbf{Dens.}     & \textbf{Div.} \\ \hline
 Bibtex 	& BTX   &  7395	& 1836	& 159	& 2.402	    & 0.015	& 0.386 \\ \hline
Birds           & BRD   & 645   & 260   & 19    & 1.014     & 0.053     & 0.206 \\ \hline
CAL500 	        & CAL   & 502	& 68	& 174	& 26.044    & 0.150	& 1.000	\\ \hline
CHD\_49         & CHD   & 555   & 49    & 6     & 2.580     & 0.430     & 0.531 \\ \hline
Enron           & ENR   & 1702  & 1001  & 53    & 3.378     & 0.064     & 0.442  \\ \hline
Flags           & FLG   & 194   & 19    & 7     & 3.392     & 0.485     & 0.422 \\ \hline
Genbase         & GBS   & 662   & 1186  & 27    & 1.252     & 0.046     & 0.048 \\ \hline
GpositivePseAAC & GPP   & 519   & 440   & 4     & 1.008     & 0.252     & 0.438 \\ \hline
Medical         & MED   & 978   & 1449  & 45    & 1.245     & 0.028     & 0.096 \\ \hline
PlantPseAAC 	& PPA   & 978	& 440	& 12	& 1.079	    & 0.090	& 0.033 \\ \hline
Scene           & SCN   & 2407  & 294   & 6     & 1.074     & 0.179     & 0.234 \\ \hline
VirusPseAAC 	& VPA   & 207	& 440	& 6	& 1.217	    & 0.203	& 0.266 \\ \hline
Water-quality   & WQT   & 1060  & 16    & 14    & 5.073     & 0.362     & 0.778  \\ \hline
Yeast           & YST   & 2417  & 103   & 14    & 4.237     & 0.303     & 0.082 \\ \hline
\end{tabular}
\end{center}
\label{table-datasets}
\end{table}

The two evolutionary \emph{search methods} (i.e., Auto-MEKA$_{GGP}$ and Auto-MEKA$_{spGGP}$) were run with 80 individuals evolved considering a time budget of five hours, tournament selection of size two, elitism of one individual, and crossover and mutation probabilities of 0.8 and 0.2, respectively. For these two methods, the learning and validation sets are also resampled from the training set every five generations in order to avoid overfitting. Additionally, we use time and memory budgets for each MLC algorithm (generated by the \emph{search methods}) of three minutes and 2GB of RAM, respectively. If any MLC algorithm reaches these budgets, it is assigned the lowest fitness, i.e., zero. Furthermore, the following convergence criterion is considered: at each iteration, we check if the best individual has remained the same for over five generations and the \emph{search method} has run for at least 20 generations. If this happens, we restart the evolutionary process with another pseudo-random seed.

In the case of Auto-MEKA$_{spGGP}$, as we have eight species, we specify 10 individuals for each species. We define Auto-MEKA$_{spGGP}$'s convergence criteria for each species individually. We set the intra-species and inter-species crossover probabilities for Auto-MEKA$_{spGGP}$ as 0.5 and 0.5, respectively. On the other hand, Auto-MEKA$_{BO}$ has kept only the time and memory budgets -- i.e., five hours of run for its respective \emph{search method}, and three minutes and 2GB of time and memory budgets for each produced MLC algorithm, respectively. As in the evolutionary methods, the MLC algorithms that reach time and memory budgets are assigned a fitness of zero. 

In order to be fair in the comparisons with the evolutionary methods, we set the value of $p$ equal to 80 for both Auto-MEKA$_{RS}$ and Auto-MEKA$_{GS}$, i.e., the methods that implement random search and greedy search, respectively.

All experiments were run using a stratified five-fold cross-valida-tion procedure \citep{Sechidis2011}. This section shows three measures considered when evaluating the results in terms of classification quality: hamming loss (HL), ranking loss (RL), and the general measure we defined as the fitness/quality criteria (see Equation \ref{eq})\footnote{We did not present and analyze the results of exact match and $F_1$ Macro averaged by label due to their similar predictive performances to HL, RL and/or fitness measures.}. Given the average values -- based on the 14 datasets -- for all methods on a particular measure, results are evaluated using an adapted Friedman test followed by a Nemenyi post-hoc test with the usual significance level of 0.05 \cite{Demvsar2006}.

\subsection{Experimental Results} \label{exp_results}

Table~\ref{measures-table} shows the average values, the average ranks and the final statistical analysis for HL, RL and fitness measures, respectively.

\begin{table}[!htbp]
\caption{Comparison of the hamming loss (to be minimized), ranking loss (to be minimized) and fitness (to be maximized) obtained by all \emph{search methods} in the test set for the three designed \emph{search spaces} with five hours of execution.}
\scriptsize
\setlength\tabcolsep{2.8pt}
\begin{center}
\begin{tabular}{|l|l|l|c|c|c|c|}
\hline
\multicolumn{1}{|c|}{\textbf{\makecell[c]{Search \\ Space}}} & \multicolumn{1}{c|}{\textbf{\makecell[c]{Evaluated \\ Result}}} & \textbf{spGGP} & \textbf{GGP} & \textbf{BO} & \textbf{RS} & \textbf{GS} \\ \hline
\multicolumn{7}{|c|}{\textbf{Hamming Loss (HL)}} \\ \hline
\multicolumn{ 1}{|l|}{\textbf{Small}} & \textbf{Avg. Values} & 0.135 & \textbf{0.134} & 0.208 & 0.137 & 0.135 \\ 
 \multicolumn{ 1}{|l|}{} & \textbf{Avg. Ranks} & 3.143 & \textbf{2.250} & 3.714 & 3.107 & 2.786 \\ 
 \multicolumn{ 1}{|l|}{} & \textbf{Stat. Comparison} & \multicolumn{ 5}{c|}{no differences among all methods} \\ \cline{ 2- 7}
\rowcolor{gray!20}
 \multicolumn{ 1}{|l|}{\textbf{Medium}} & \textbf{Avg. Values} & 0.157 & 0.136 & \textbf{0.134} & 0.139 & 0.142 \\ 
\rowcolor{gray!20}
 \multicolumn{ 1}{|l|}{} & \textbf{Avg. Ranks} & 4.036 & \textbf{2.251} & 2.607 & 3.00 & 3.107 \\ 
\rowcolor{gray!20}
 \multicolumn{ 1}{|l|}{} & \textbf{Stat. Comparison} & \multicolumn{ 5}{c|}{\{GGP\} $\succ$ \{spGGP\}, no differences among the others} \\ \cline{ 2- 7}
\rowcolor{gray!50}
 \multicolumn{ 1}{|l|}{\textbf{Large}} & \textbf{Avg. Values} & 0.137 & {0.134} & \textbf{0.130} & 0.143 & 0.139 \\ 
\rowcolor{gray!50}
 \multicolumn{ 1}{|l|}{} & \textbf{Avg. Ranks} & 3.214 & 2.571 & \textbf{2.07} & 4.00 & 3.142 \\ 
\rowcolor{gray!50}
 \multicolumn{ 1}{|l|}{} & \textbf{Stat. Comparison} & \multicolumn{ 5}{c|}{\{BO\} $\succ$ \{RS\}, no differences among the others} \\ \hline
\multicolumn{7}{|c|}{\textbf{Ranking Loss (RL)}} \\ \hline
\multicolumn{ 1}{|l|}{\textbf{Small}} & \textbf{Avg. Values} & \textbf{0.153} & 0.161 & 0.210 & 0.167 & 0.167 \\ 
 \multicolumn{ 1}{|l|}{} & \textbf{Avg. Ranks} & 2.321 & \textbf{2.214} & 3.750 & 3.429 & 3.286 \\ 
 \multicolumn{ 1}{|l|}{} & \textbf{Stat. Comparison} & \multicolumn{ 5}{c|}{no differences among all methods} \\ \cline{ 2- 7}
\rowcolor{gray!20}
 \multicolumn{ 1}{|l|}{\textbf{Medium}} & \textbf{Avg. Values} & \textbf{0.146} & 0.150 & 0.152 & 0.156 & 0.148 \\ 
\rowcolor{gray!20}
 \multicolumn{ 1}{|l|}{} & \textbf{Avg. Ranks} & \textbf{2.679} & 2.821 & 3.250 & 3.321 & 2.929 \\ 
\rowcolor{gray!20}
 \multicolumn{ 1}{|l|}{} & \textbf{Stat. Comparison} & \multicolumn{ 5}{c|}{no differences among all methods} \\ \cline{ 2- 7}
\rowcolor{gray!50}
 \multicolumn{ 1}{|l|}{\textbf{Large}} & \textbf{Avg. Values} & 0.147 & \textbf{0.135} & 0.149 & 0.157 & 0.140 \\ 
\rowcolor{gray!50}
 \multicolumn{ 1}{|l|}{} & \textbf{Avg. Ranks} & 2.821 & \textbf{2.536} & 2.786 & 4.071 & 2.786 \\ 
\rowcolor{gray!50}
 \multicolumn{ 1}{|l|}{} & \textbf{Stat. Comparison} & \multicolumn{ 5}{c|}{no differences among all methods} \\ \hline
\multicolumn{7}{|c|}{\textbf{Fitness}} \\ \hline
\multicolumn{ 1}{|l|}{\textbf{Small}} & \textbf{Avg. Values} & 0.626 & \textbf{0.631} & 0.590 & 0.624 & 0.626 \\ 
 \multicolumn{ 1}{|l|}{} & \textbf{Avg. Ranks} & 2.679 & \textbf{1.964} & 3.857 & 3.535 & 2.964 \\ 
 \multicolumn{ 1}{|l|}{} & \textbf{Stat. Comparison} & \multicolumn{ 5}{c|}{\{GGP\} $\succ$ \{BO\}, no differences among the others} \\ \cline{ 2- 7}
\rowcolor{gray!20}
 \multicolumn{ 1}{|l|}{\textbf{Medium}} & \textbf{Avg. Values} & 0.634 & 0.639 & 0.626 & 0.629 & \textbf{0.645} \\ 
\rowcolor{gray!20}
 \multicolumn{ 1}{|l|}{} & \textbf{Avg. Ranks} & 3.107 & \textbf{2.643} & 2.929 & 3.429 & 2.893 \\ 
\rowcolor{gray!20}
 \multicolumn{ 1}{|l|}{} & \textbf{Stat. Comparison} & \multicolumn{ 5}{c|}{no differences for all methods} \\ \cline{ 2- 7}
\rowcolor{gray!50}
 \multicolumn{ 1}{|l|}{\textbf{Large}} & \textbf{Avg. Values} & 0.632 & \textbf{0.635} & \textbf{0.635} & 0.626 & 0.632 \\ 
\rowcolor{gray!50}
 \multicolumn{ 1}{|l|}{} & \textbf{Avg. Ranks} & 2.607 & \textbf{2.429} & 2.571 & 4.286 & 3.107 \\ 
\rowcolor{gray!50}
 \multicolumn{ 1}{|l|}{} & \textbf{Stat. Comparison} & \multicolumn{ 5}{c|}{\{spGGP, GGP, BO\} $\succ$ \{RS\}, no differences among the others} \\ \hline
\end{tabular}
\end{center}
\label{measures-table}
\end{table}

The results for the HL measure with five hours, in Table~\ref{measures-table}, showed that Auto-MEKA$_{GGP}$ had the best average values and ranks in the \emph{search space} Small, but it had only the best average rank in the \emph{search space} Medium.  In addition, Auto-MEKA$_{BO}$ obtained the best average value and Auto-MEKA$_{GGP}$ was statistically better than Auto-MEKA$_{spGGP}$ in the \emph{search space} Medium. Nonetheless, there is no indication of statistical difference for the other cases of \emph{search space} Medium, neither for any cases of Small. On the other hand, when considering the \emph{search space} Large, Auto-MEKA$_{BO}$ had the best average value and the best average rank. By looking at the statistical results of HL in Table~\ref{measures-table}, we can observe that only Auto-MEKA$_{BO}$ was able to significantly outperform Auto-MEKA$_{RS}$ for the \emph{search space} Large. In the other cases, we do not see an indication of the \emph{search methods} to trade-off well between exploration and exploitation. Therefore, based on these HL results, we can conclude  that the size of the \emph{search space} has a strong influence on the performance of the \emph{search method}.

In addition, Table~\ref{measures-table} summarizes the results for the evaluation measure from a different evaluation context, i.e., RL. Whilst Auto-MEKA$_{spGGP}$ was the \emph{search method} with the best average value in the \emph{search space} Small, $\text{Auto-MEKA}_{GGP}$ achieved the best average rank in this scenario. Distinctly, $\text{Auto-MEKA}_{spGGP}$ selected and configured MLC algorithms in a way that produced the best average value and rank in the \emph{search space} Medium. For the \emph{search space} Large, $\text{Auto-MEKA}_{GGP}$ was the \emph{search method} with the best average value and the best average rank.

The results of RL did not present any evidence of statistical significance. Hence, the \emph{search methods} did not differ from each other. As this metric comes from another context, we would like to understand why it presented such a flat result for all \emph{search spaces} and \emph{search methods}. We believe that the RL measure is very conservative and it is not sensitive to different choices of multi-label classifiers. As it only penalizes reversed pairs of labels into the ranking and it does not take into account the label-pair depth into the ranking to penalize, this can make this measure not good enough to be used in isolation to evaluate MLC algorithms. In future work, it might be interesting to evaluate whether this measure is appropriate to be part of our study or whether we should consider a rank-based measure that takes into account the depth~of~the~ranking.

HL and RL are measures that compose the fitness/quality function. We also analyze the results of the fitness measure to have a better assessment of the results of \emph{search methods}. In the last part of Table~\ref{measures-table}, we show the results for this measure. With respect to the \emph{search space} Small, Auto-MEKA$_{GGP}$ presented the best average value and also the best average rank. Besides, we found statistical evidence that Auto-MEKA$_{GGP}$ has better results than Auto-MEKA$_{BO}$ in this \emph{search space}. In the \emph{search space} Medium, we observe that Auto-MEKA$_{GGP}$ produced the best average ranking and Auto-MEKA$_{GS}$ presented the best average value. Finally, in the comparison regarding the \emph{search space} Large, Auto-MEKA$_{spGGP}$, Auto-MEKA$_{GGP}$ and Auto-MEKA$_{BO}$ achieved significantly better results than Auto-MEKA$_{RS}$, showing their capabilities to handle enormous \emph{search spaces} -- when giving enough time for them to proceed with their searches. Apart from the statistical results, we can observe in Table~\ref{measures-table} that Auto-MEKA$_{GGP}$ also reached the best average value and rank within five hours of running. Furthermore, Auto-MEKA$_{BO}$ had even results to Auto-MEKA$_{GGP}$ in terms of the average value. 

We can now provide overall conclusions based on the results of this section, specially on the fitness measure. We believe that the size of the \emph{search spaces} explored/exploited by the \emph{search methods} influenced the accuracy of the MLC predictions. We understand that, for smaller \emph{search spaces} (i.e., Small and Medium), the \emph{search methods} find it easier to proceed with their searches and, hence, their results become broadly similar among each other. When we increase the \emph{search space} to Large, only those with robust search mechanisms could deal better with the trade-off between exploration and exploration, making them achieve competitive results.

However, we believe the robust methods (i.e., those based on the evolutionary and Bayesian optimization frameworks) can still improve their final predictive performances as they could not beat the greedy \emph{search method}, a pure-exploitation method. They also face challenges to beat the random \emph{search method} in some of the analyzed cases. Thus, these \emph{search methods} still could not satisfactorily balance between exploration and exploitation. In our perspective, this would occur if they could beat simultaneously Auto-MEKA$_{RS}$ and Auto-MEKA$_{GS}$. For the \emph{search spaces} Small and Medium, this result is more understandable. The smaller the \emph{search space}, the easier it is to perform the search on it. This leads to better results for Auto-MEKA$_{RS}$ and Auto-MEKA$_{GS}$ in such a way that the other \emph{search methods} were not statistically different from both of them. For bigger \emph{search spaces} (i.e., Large), this should be the opposite. As they have robust search mechanisms, they should obtain statistically better results against pure-exploration and pure-exploitation \emph{search methods}.
\subsection{Analysis of the Diversity of the Selected Algorithms} \label{selected_algorithms}

We also analyzed the diversity of the MLC algorithms and meta-algorithms selected by the five AutoMLC \emph{search methods}. We focus only on the selected MLC algorithms and meta-algorithms (which are the ``macro-components''), and not on their selected SLC algorithms and hyper-parameter settings (the ``micro-components''), to simplify our analysis. 

It is important to emphasize that, by analyzing the MLC and SLC algorithms and meta-algorithms selected by all \emph{search methods} in each \emph{search space}, we can better understand the results of Table \ref{measures-table}. This would give an idea of how the choice of an MLC algorithm influences the performance of the \emph{search methods}. However, for the sake of simplicity, we perform this analysis only for the \emph{search space} Large. 

 \begin{figure*}[!ht]
     \centering
     \begin{subfigure}[t]{0.305\textwidth}
         \centering
         \includegraphics[scale=0.22147065325001998752351025]{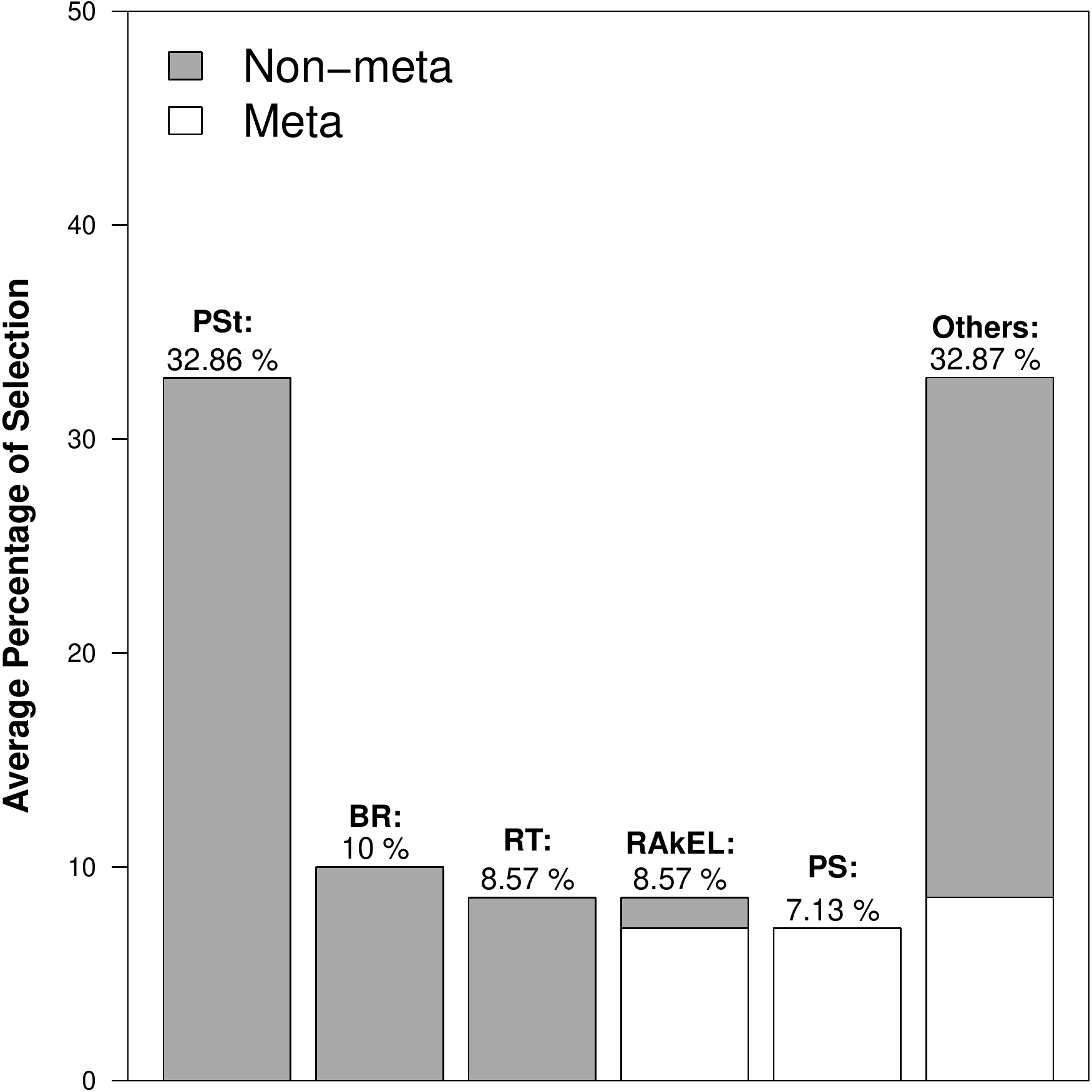}
         \caption{Auto-MEKA$_{spGGP}$.}
         \label{barplot-spggp_mlc}
     \end{subfigure}
     ~     
     \begin{subfigure}[t]{0.305\textwidth}
         \centering
         \includegraphics[scale=0.22147065325001998752351025]{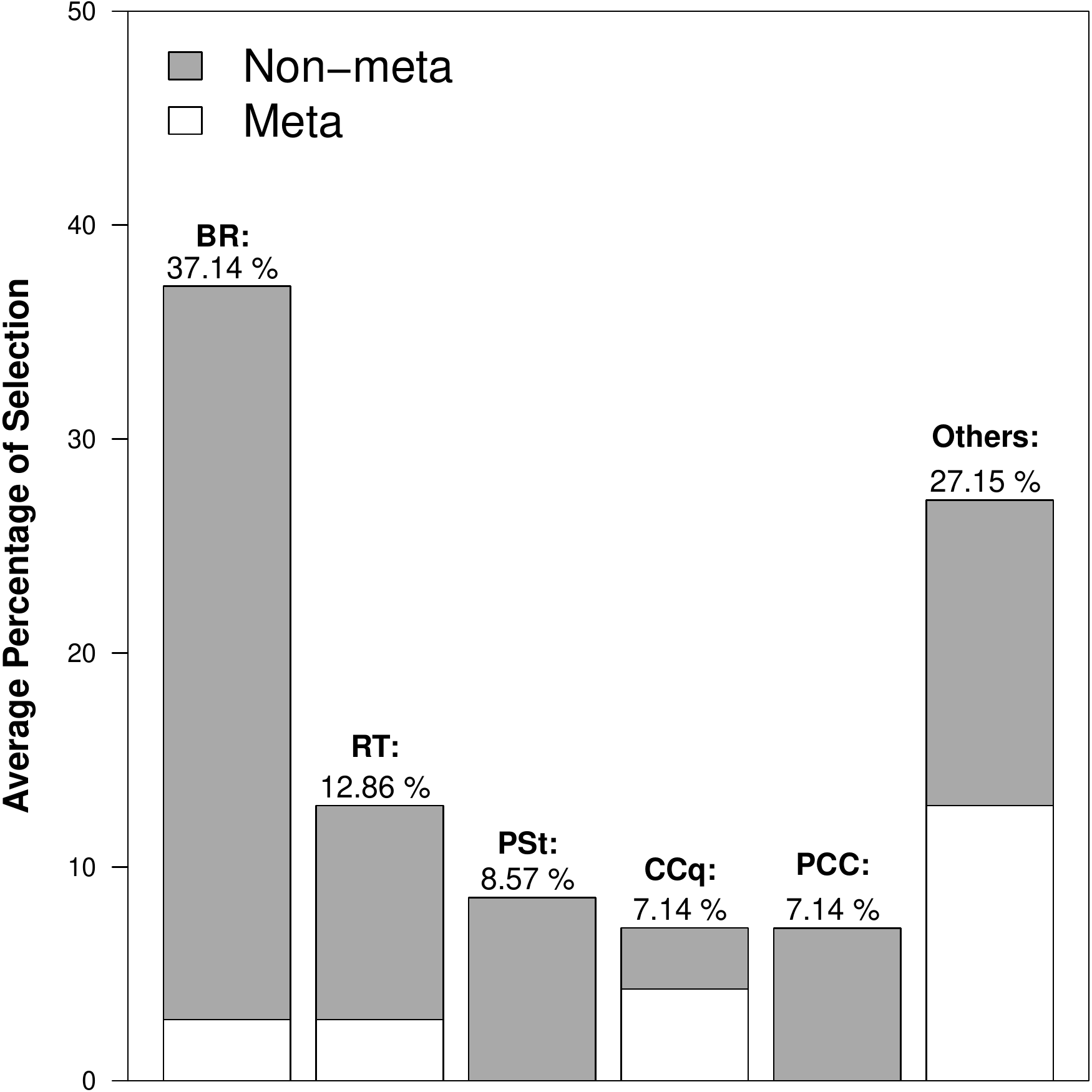}
         \caption{Auto-MEKA$_{GGP}$.}
         \label{barplot-ggp_mlc}
     \end{subfigure}%
     ~ 
     \begin{subfigure}[t]{0.305\textwidth}
         \centering
         \includegraphics[scale=0.22147065325001998752351025]{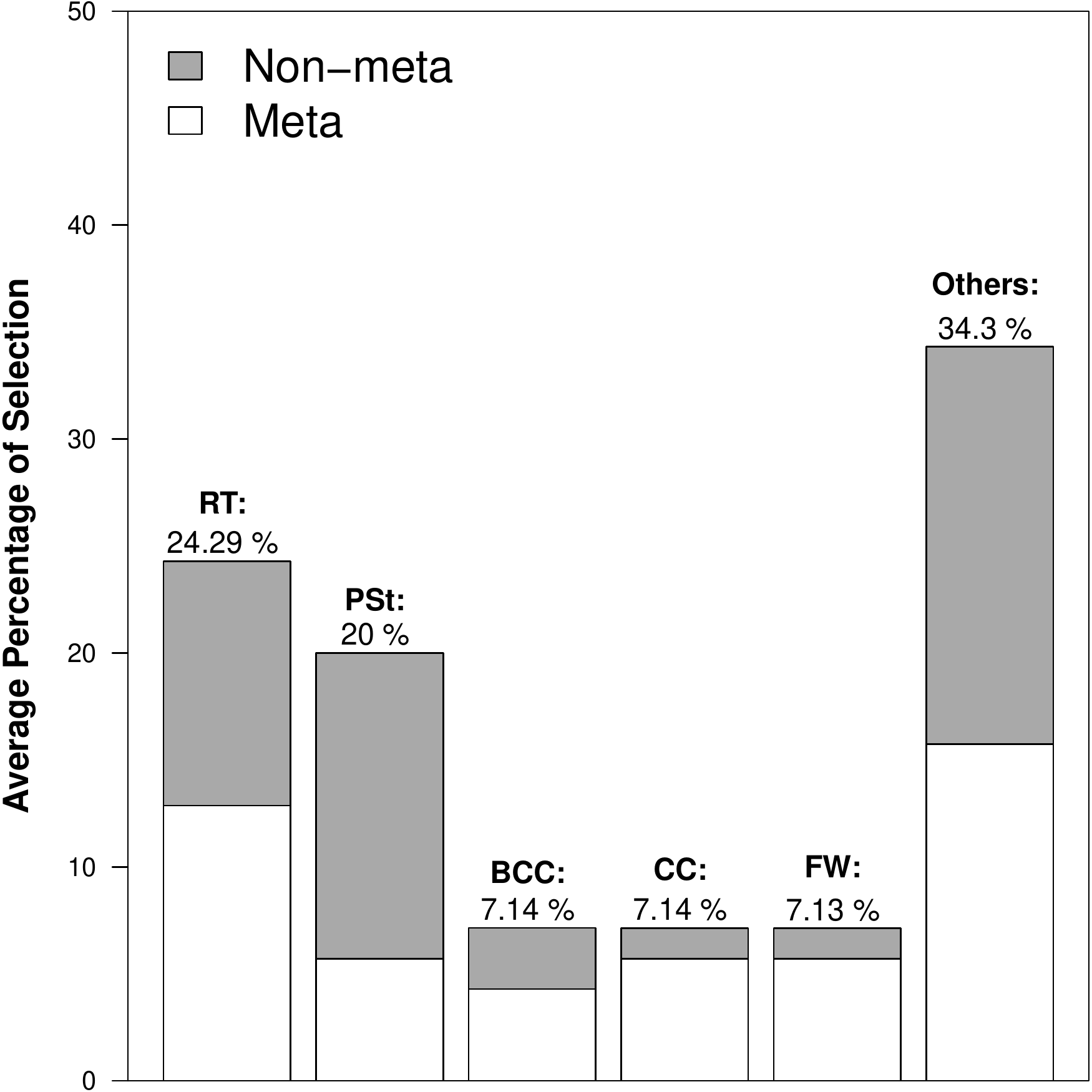}
         \caption{Auto-MEKA$_{BO}$.}
         \label{barplot-bo_mlc}
     \end{subfigure}%
     
     \begin{subfigure}[t]{0.305\textwidth}
         \centering
         \includegraphics[scale=0.22147065325001998752351025]{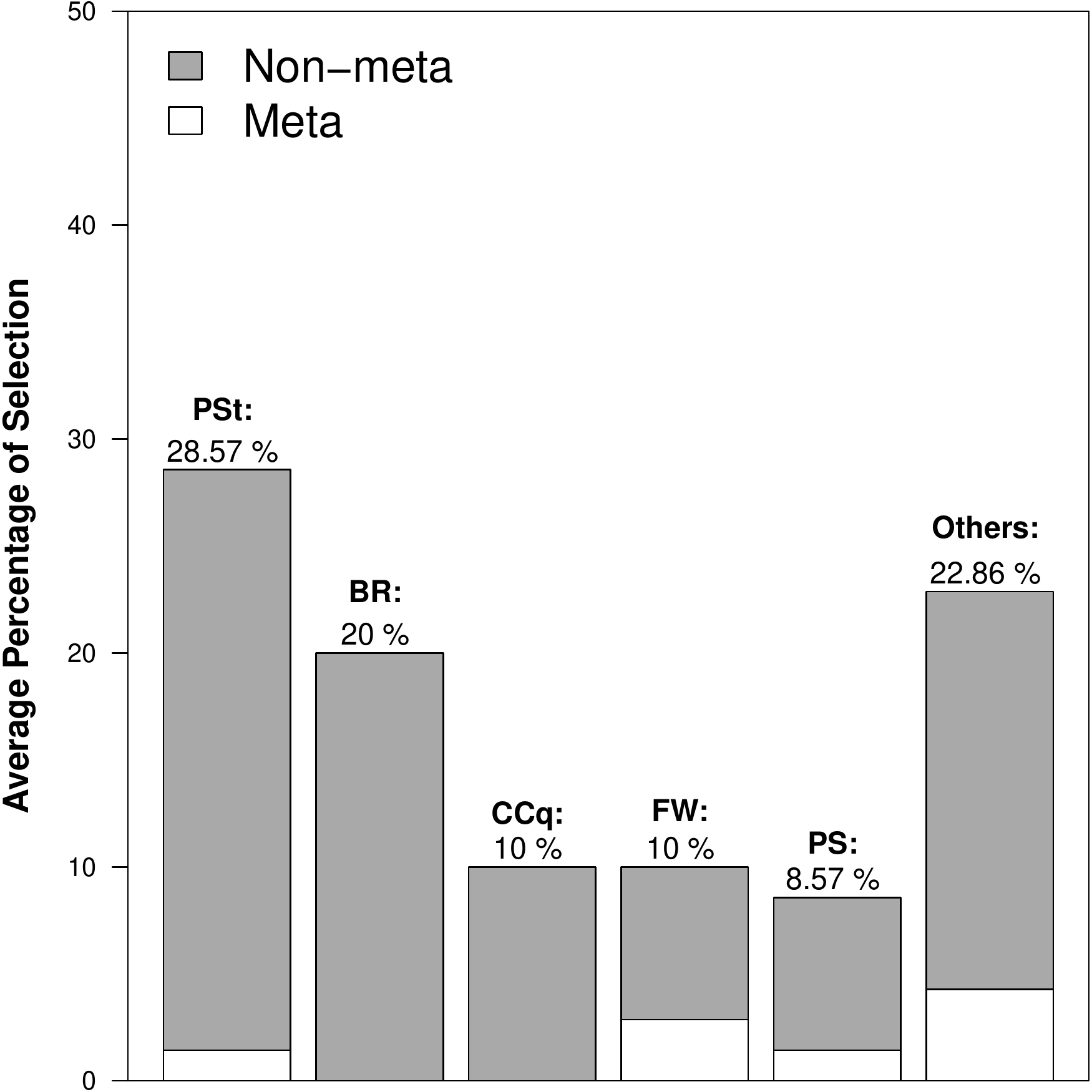}
         \caption{Auto-MEKA$_{RS}$.}
         \label{barplot-rs_mlc}
     \end{subfigure}
     ~
     \begin{subfigure}[t]{0.305\textwidth}
         \centering
         \includegraphics[scale=0.22147065325001998752351025]{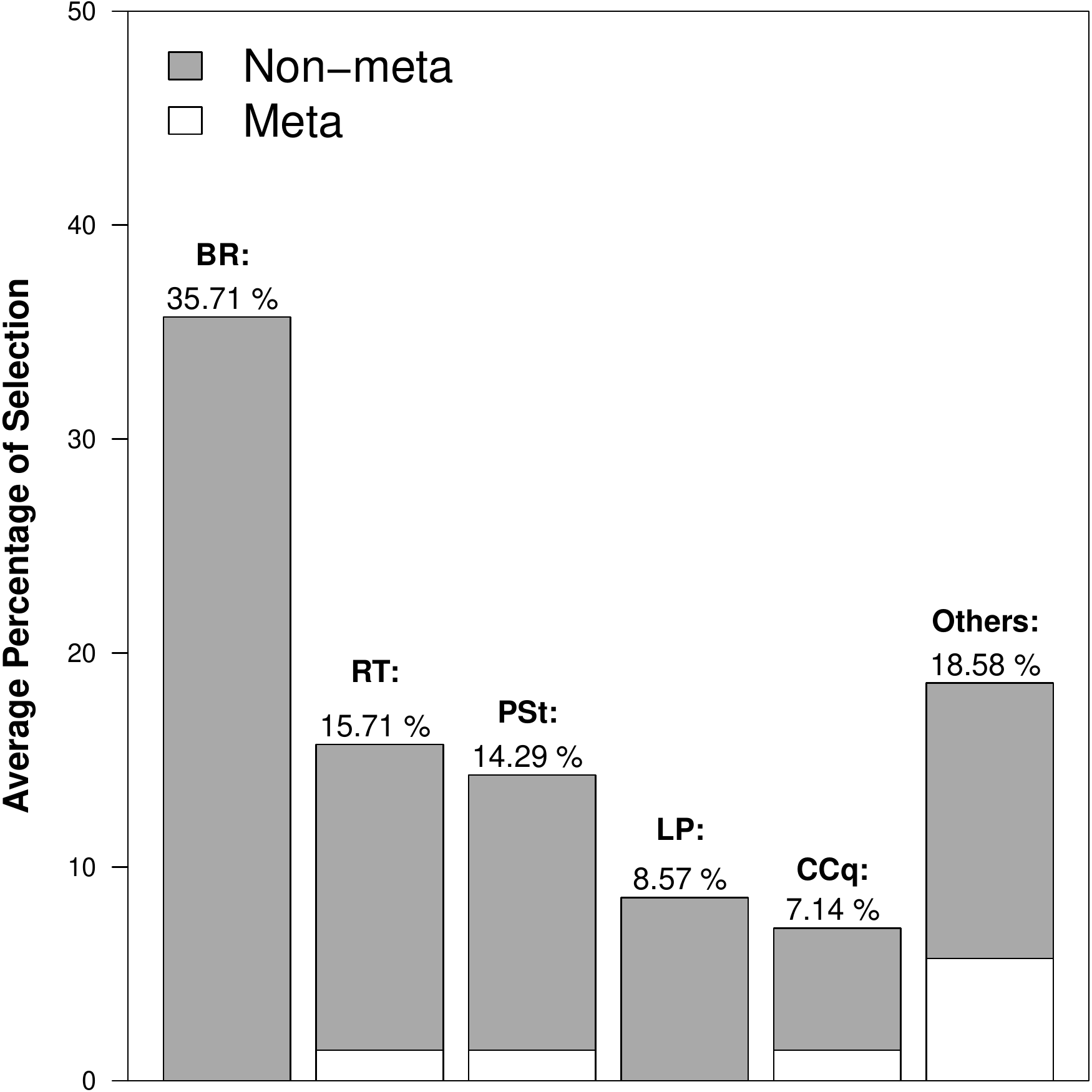}
         \caption{Auto-MEKA$_{GS}$.}
         \label{barplot-bs_mlc}
     \end{subfigure}%
     
     \caption{Bar plots for the algorithms' selection frequencies at the MLC level over 70 runs.}
 \end{figure*}

We present in Figures \ref{barplot-ggp_mlc} through \ref{barplot-bs_mlc} the bar plots to analyze the relative frequency of selection of MLC algorithms for the  AutoMLC \emph{search methods}. In these figures we have, for each MLC algorithm, a bar representing the average relative frequency of selection of an algorithm type over all the {70} runs: {14} datasets times five independent runs per dataset (five cross-validation folds times one run per fold). We consider two cases: (i) when the traditional MLC algorithm is solely selected; (ii) when the traditional MLC algorithm is selected together with a MLC meta-algorithm. To emphasize these two cases, the bar for each traditional MLC algorithm is divided into two parts, with sizes proportional to the relative frequency of selection as a standalone algorithm (in gray color) and the relative frequency of selection as part of a meta-algorithm (in white color).

Considering this information, BR, PSt and RT were the traditional MLC algorithms most frequently selected by all AutoMLC \emph{search methods} in the \emph{search space} Large. BR was chosen, on average, in 21.43\% of all runs for all methods. PSt and RT, in turn, were selected on average in 20.66\% and 13.43\% of all runs for the five evaluated \emph{search methods}, respectively. Nevertheless, some of these MLC algorithms were not so present in the selections performed by the \emph{search methods}. For instance, BR and RT were not frequently chosen by Auto-MEKA$_{BO}$ and Auto-MEKA$_{RS}$. This partially shows the differences in the selection and configuration of the AutoMLC \emph{search methods}, although most of them had similar algorithms at the top five regarding the ranking of selection.

We can also justify the performance of the methods based on their selection at the MLC meta level. For example, Auto-MEKA$_{GGP}$ achieved the best results for the \emph{search space} Large in terms of the average value and rank based on fitness, which is the measure we use to decide (for all methods) what algorithm is the most appropriate. By looking at Auto-MEKA$_{GGP}$'s selection at the MLC meta level we can understand why this happened. Auto-MEKA$_{GGP}$ and Auto-MEKA$_{spGGP}$ have chosen these MLC meta-algorithms with a low relative frequency (22.86\% for both methods). Therefore, the complexity of the final solution turned them into better options for AutoMLC in the MLC context when contrasted to Auto-MEKA$_{BO}$, which selected meta-algorithms in 50\% of the cases. However, their level of selection of MLC meta-algorithms is still high when compared to Auto-MEKA$_{RS}$ and Auto-MEKA$_{GS}$, which selected MLC meta-algorithms in only 10\% of the cases. This might be the reason for the competitiveness of Auto-MEKA$_{RS}$ and Auto-MEKA$_{GS}$.

One test that might be interesting is to remove the MLC meta-algorithms from the \emph{search space}, and reexecute the \emph{search methods}. This could show us whether or not the learned model is more likely to overfit on the training set when we select very complex combinations of base and meta-algorithms. We did that in the \emph{search spaces} Small and Medium, but they do not include all traditional MLC algorithms as the \emph{search space} Large does.
\section{Conclusions}
\label{sec:conclusions}

This paper presented an overall comparison among five AutoML \emph{search methods} in the context of multi-label classification -- i.e., Auto-MEKA$_{GGP}$, Auto-MEKA$_{spGGP}$, Auto-MEKA$_{BO}$, Auto-MEKA$_{RS}$ and Auto-MEKA$_{GS}$. To perform this assessment, the \emph{search methods} were run in 14 MLC datasets with the same execution time budget (i.e., five hours) and in three designed \emph{search spaces}. 

The experimental results indicate that Auto-MEKA$_{GGP}$ is so far the best \emph{search method} as it yields the best predictive results. Besides, it is the only method to be statistically better than Auto-MEKA$_{spGGP}$, Auto-MEKA$_{BO}$ and Auto-MEKA$_{RS}$ in different cases.

However, we expected that methods with robust search mechanisms (e.g., Auto-MEKA$_{spGGP}$, Auto-MEKA$_{BO}$ and Auto-MEKA$_{GGP}$) could balance better between exploration and exploitation. This limitation made these methods not being able to produce statistically better results than Auto-MEKA$_{RS}$ and Auto-MEKA$_{GS}$, which are pure-exploration and pure-exploitation methods, respectively. 

We also observed that the size of the \emph{search space} is a crucial issue for the AutoML methods' behavior. Thus, as a first future work, we intend to better understand the trade-off between parsimony and sufficiency \cite{Banzhaf1998}. In other words, we would like to investigate which algorithms should be included to or excluded from the \emph{search spaces}, in order to keep good (combinations of) learning algorithms.

In addition, as two out of the five measures (i.e., FM and RL) yielded flat results, we need to understand how neutral the \emph{search spaces} are in terms of the chosen MLC performance measures \cite{Pitzer2012,Malan2013}. This would help us to have insights to propose efficient methods or enhancements to these \emph{search spaces}. 

Furthermore, we expect to test other quality criteria to discover appropriate MLC algorithms configurations for a given dataset of interest. This may include finding other relevant performance measures or to set new weights for the current measures that compose the proposed fitness metric.

Finally, we also plan to include into the proposed \emph{search methods} a bi-level optimization approach \cite{Talbi2013} to diminish the hardness of the search in huge \emph{search spaces}. Fundamentally, this would mean to select the learning algorithms in the first place and only configure their hyper-parameters in a second step of the search procedure.

\section*{Acknowledgements}

The authors would like to thank FAPEMIG (through the grant no. CEX-PPM-00098-17), MPMG (through the project Analytical Capabilities), CNPq (through the grant no. 310833/2019-1), CAPES, MCTIC/RNP (through the grant no. 51119) and H2020 (through the grant no. 777154) for their partial financial support. 

\bibliographystyle{ACM-Reference-Format}
\bibliography{bibfile.bib} 

\end{document}